\PassOptionsToPackage{table}{xcolor}

\documentclass[10pt,twocolumn,letterpaper]{article}

\usepackage{cvpr}              

\usepackage[dvipsnames]{xcolor}

\definecolor{ph-purple}{RGB}{129, 39, 232}
\definecolor{ph-blue}{RGB}{5, 131, 227}
\definecolor{ph-gray}{rgb}{0.5, 0.5, 0.5}
\definecolor{ph-orange}{RGB}{227, 127, 5}
\definecolor{ph-green}{RGB}{0, 135, 124}
\definecolor{ph-yellow}{RGB}{235, 201, 52}
\definecolor{ph-light-green}{RGB}{181, 209, 21}
\definecolor{ph-red}{RGB}{250, 101, 60}
\definecolor{ph-indigo}{RGB}{75, 0, 130}

\colorlet{ph-orange-light}{ph-orange!70}
\colorlet{ph-blue-light}{ph-blue!70}
\colorlet{ph-purple-light}{ph-purple!70}
\colorlet{ph-green-light}{ph-green!70}
\definecolor{ph-light-gray}{rgb}{0.75, 0.75, 0.75}

\newcommand{\IGNORE}[1]{}
\newcommand{\mparagraph}[1]{\noindent\textbf{#1}}

\newcommand{\reffig}[1]{Fig.~\ref{#1}}
\newcommand{\reftab}[1]{Tab.~\ref{#1}}
\newcommand{\refsec}[1]{Sec.~\ref{#1}}

\newcommand{\anidress}{\emph{AniDress}}

\newcommand{\cmark}{\color{green}\ding{51}}%
\newcommand{\xmark}{\color{red}\ding{55}}%

\definecolor{cvprblue}{rgb}{0.21,0.49,0.74}
\usepackage[pagebackref,breaklinks,colorlinks,citecolor=cvprblue]{hyperref}
\usepackage{multirow}
\usepackage{tabularx}
\usepackage{pifont}

\title{\anidress: Animatable Loose-Dressed Avatars from Sparse Views Using Garment Rigging Models}

\renewcommand{\thefootnote}{\fnsymbol{footnote}}
\author{
Beijia Chen\footnotemark[1]
\quad Yuefan Shen\footnotemark[1]
\quad Qing Shuai
\quad Xiaowei Zhou
\quad Kun Zhou
\quad Youyi Zheng$^{\dag}$
\\
\\
State Key Lab of CAD\&CG, Zhejiang University
}

\begin{document}
\maketitle
\thispagestyle{plain}
\pagestyle{plain}
\begin{abstract}
\vspace{-0.44cm}
Recent communities have seen significant progress in building photo-realistic animatable avatars from sparse multi-view videos. 
However, current workflows struggle to render realistic garment dynamics for loose-fitting characters as they predominantly rely on naked body models for human modeling while leaving the garment part un-modeled. 
This is mainly due to that the deformations yielded by loose garments are highly non-rigid, and capturing such deformations often requires dense views as supervision. 
In this paper, we introduce AniDress, a novel method for generating animatable human avatars in loose clothes using very sparse multi-view videos (4-8 in our setting). 
To allow the capturing and appearance learning of loose garments in such a situation, we employ a virtual bone-based garment rigging model obtained from physics-based simulation data.
Such a model allows us to capture and render complex garment dynamics through a set of low-dimensional bone transformations.
Technically, we develop a novel method for estimating temporal coherent garment dynamics from a sparse multi-view video. 
To build a realistic rendering for unseen garment status using coarse estimations, a pose-driven deformable neural radiance field conditioned on both body and garment motions is introduced, providing explicit control of both parts. At test time, the new garment poses can be captured from unseen situations, derived from a physics-based or neural network-based simulator to drive unseen garment dynamics.
To evaluate our approach, we create a multi-view dataset that captures loose-dressed performers with diverse motions.
Experiments show that our method is able to render natural garment dynamics that deviate highly from the body and generalize well to both unseen views and poses, surpassing the performance of existing methods.
The code and data will be publicly available.
\vspace{-0.66cm}

\end{abstract}    

\footnotetext[1]{Equal Contributions, $^{\dag}$Corresponding Author.}
\renewcommand{\thefootnote}{\arabic{footnote}}

\section{Introduction}
\label{sec:intro}

\begin{figure}[t!]
    \centering
    \includegraphics[width=\linewidth]{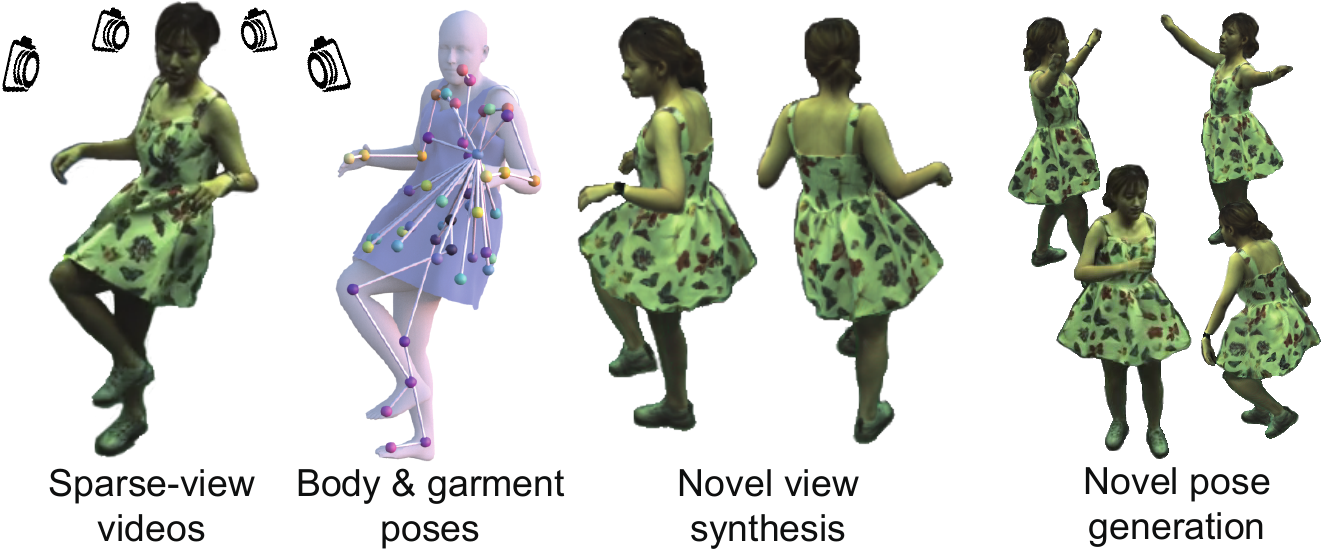}
    \caption{Given a sparse multi-view video with a loose-dressed performer, we estimate both body and garment poses aided by a garment rigging model. Then, a pose-driven neural radiance field is optimized to fit the video. At test time, our method can synthesize plausible body and garment motions from novel views. In this case, we use body motions from AMASS~\cite{AMASS:2019} and garment poses from physics-based simulation for novel pose synthesis.}
    \label{fig:teaser}
\end{figure}

Building animatable clothed human avatars has long been a challenging task in VR and AR. Recent neural rendering-based works \cite{peng2021neural, peng2021animatable, liu2021neural, noguchi2021neural, zheng2022structured, li2022tava, wang2022arah, Zheng2023AvatarRex} enable us to learn animatable characters directly from sparse multi-view videos (usually consists of 300-400 frames) and produce photo-realistic rendering. Generally, these methods first capture the body movements using template-based estimators \cite{easymocap} and then draft the estimated deformations using a NeRF-based rendering model for appearance learning. At test time, these methods generate novel animations driven by unseen body poses.


Though appealing results have been achieved, these methods fail when applied to loose-fitting subjects. 
This is due to the fact that loose garments such as dresses and skirts, exhibiting complex non-rigid deformations such as swing and sliding effects, can own different movement patterns from body~\cite{ma2021power, ma2022neural, xiang2021modeling, xiang2022dressing, pan2022predicting}.
Using 3D body poses alone for capturing cross-frame correspondences is inaccurate, producing blurry renderings at training time.
Moreover, these approaches generate novel animations driven by body pose only, failing to model the complex interplay between body and cloth dynamics at test time. 


To address the above challenges, several methods have been introduced. HDHumans \cite{habermann2023hdhumans} exploits deformation graphs \cite{sumner2007embedded} for modeling garment movements. However, their method requires additional annotations of ground-truth texture for each frame during training. Moreover, they learn the body-to-garment mapping directly from multi-view videos of limited length, resulting in stiff animations at test time. Other works ~\cite{xiang2021modeling, xiang2022dressing} track and reconstruct accurate garment meshes at the texture-aligned level for learning realistic rendering models. Novel garment statuses upon unseen body poses are obtained from physics-based simulation to render garment animations at test time. However, they rely on expensive dense views of cameras and cannot be applied to sparse views.

In this paper, we present \anidress, a novel solution for building generalizable loose-fitting avatars from a sparse multi-view video.
Instead of directly learning geometry animation model and appearance model from limited video data, we follow the previous work \cite{xiang2022dressing} that only learns the rendering model from multi-view videos. To enable garment dynamics capture and rendering in very sparse views, we employ a garment rigging model built from physics-based simulation (PBS) data.
Specifically, given a garment template (obtained from reconstruction), we first compute diverse garment dynamics for different body animations using PBS.
Based on these mesh examples, a virtual bone-based garment rigging model is extracted via skinning decomposition \cite{le2012smooth}.
This model encodes high-dimensional vertex deformations into a low-dimensional intrinsic garment space, empowering us with several advantages.
First, it allows us to capture garment dynamics effectively using a set of bone transformations (denoted as garment poses for brevity).
Technically, we introduce a novel method based on differentiable rendering to estimate temporally coherent garment poses from a multi-view video.
Second, the estimated garment poses, in conjunction with body poses, constitute a unified signal to drive successive rendering.
Specifically, we develop a pose-driven NeRF-based rendering method where both body and garment poses are used to deform a neural radiance field. 
Our method can generate photo-realistic avatars with controllability over the body and garment poses.
At test time, garment poses can be derived from PBS, predicted from learning-based methods, or estimated from videos, to control the rendering of garment dynamics (see \reffig{fig:teaser}).

To evaluate our approach, we create a multi-view dataset capturing performers in diverse loose garments under various motions.  Experiments show that our approach outperforms prior works in terms of both novel view and novel pose synthesis. 

In summary, our contributions are as follows:
\begin{itemize}

\item We adopt a garment rigging model built from simulation data for capturing, animating, and rendering garment dynamics. 

\item A novel method to estimate garment poses from sparse multi-view RGB videos.

\item A deformable neural radiance field conditioned on both garment and body poses, allowing for rendering high-quality body movements with natural garment dynamics.

\item A multi-view dataset that captures five loose dresses in diverse motions for evaluation and further research.

\end{itemize}
\section{Related Works}

\begin{table}[b!]
    \caption{We compare the characteristics of our method with representative existing methods. Our method is the only one that can create loose-dressed avatars from sparse RGB views.}
    \begin{tabular}{c|cccc}
    \hline
    Methods    & \begin{tabular}[c]{@{}c@{}}Sparse\\ views\end{tabular}
               & \begin{tabular}[c]{@{}c@{}}RGB\\ only\end{tabular}
               & \begin{tabular}[c]{@{}c@{}}Novel\\ pose\end{tabular}
               & \begin{tabular}[c]{@{}c@{}}Loose-\\ dressed\end{tabular} \\ \hline 
    NeuralBody~\cite{peng2021neural}       & \cmark & \cmark & \cmark & \xmark \\
    Anim-Nerf~\cite{peng2022animatable}   & \cmark & \cmark & \cmark & \xmark \\
    Uv-Volumes~\cite{chen2023uv}           & \cmark & \cmark & \cmark & \xmark \\ \hline
    Xiang \etal~\cite{xiang2022dressing}       & \xmark & \cmark & \cmark & \cmark \\
    Xiang \etal~\cite{xiang2023drivable}       & \xmark & \xmark & \cmark & \cmark \\ \hline
    Zhao \etal~\cite{zhao2022human}           & \xmark & \cmark & \xmark & \cmark \\
    HumanRF~\cite{isik2023humanrf}         & \xmark & \cmark & \xmark & \cmark \\
    FlexNeRF~\cite{jayasundara2023flexnerf} & \cmark & \cmark & \xmark & \cmark \\ \hline
    Ours       & \cmark & \cmark & \cmark & \cmark \\ \hline
    \end{tabular}
    \label{tab:charac}
\end{table}

\subsection{Clothed Body Avatars}
Pioneer works learn animatable avatars from real scans ~\cite{zhang2023closet,  ma2020learning, ma2021power, ma2022neural, saito2021scanimate, tiwari21neuralgif, chen2021snarf, li2022avatarcap}, exploring various geometry representations ranging from topology-fixed meshes \cite{ma2020learning} to topology-flexible forms such as point clouds \cite{ma2021power, ma2022neural} and implicit functions \cite{tiwari21neuralgif, palafox2021npms, mihajlovic2021leap, saito2021scanimate, chen2021snarf}.
However, these methods necessitate 3D body scans for training and additional rendering models for generating high-quality images.

Recent advances in neural rendering have facilitated learning animatable avatars directly from multi-view videos~\cite{pumarola2021d, peng2021neural, peng2021animatable, wang2022arah, li2022tava, chen2023uv}.
Pioneer work~\cite{pumarola2021d} learns a NeRF-based rendering model in canonical space and drafts a deformation network for dynamic modeling, requiring dense view cameras for supervision.
Subsequent research reduces the capture requirements from dense to sparse views \cite{peng2021neural, peng2021animatable} or even to monocular setups \cite{weng2022humannerf,  yu2023monohuman} and boosts its rendering speed from offline to real-time \cite{zhao2022human, chen2023uv, geng2023learning}.
Additionally, geometry constraints \cite{li2022tava, wang2022arah} and advanced dynamics learning schemes \cite{weng2022humannerf, weng2023personnerf, yu2023monohuman} have been investigated to develop more generalizable models for unseen poses.

\begin{figure*}[t!]
    \centering
    \includegraphics[width=\linewidth]{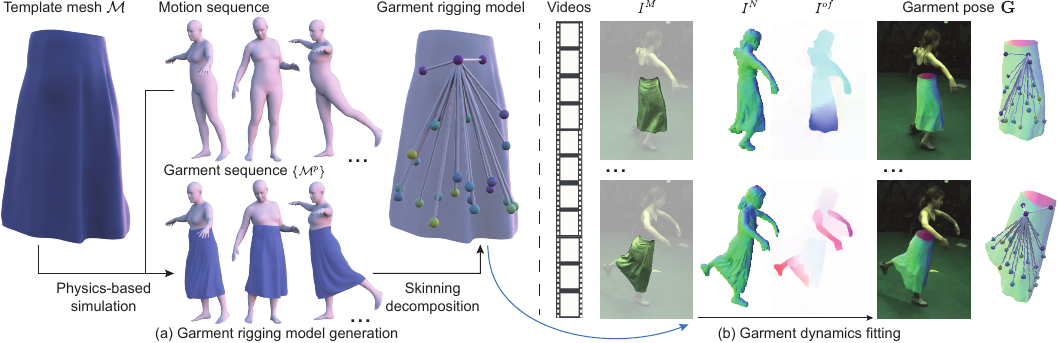}
    \caption{Overview of the procedures for building garment rigging model and capturing garment poses from a multi-view video. Starting from a template mesh $\mathcal{M}$, we run the physics-based simulation to generate diverse garment shapes $\{\mathcal{M}^p\}$, from which a garment LBS modeling is extracted via skinning decomposition. In the fitting step, we use the garment masks $I^M$, image normals $I^N$, and optical flows $I^{of}$ to estimate the garment poses at each frame.}
    \label{fig:gmlbs}
\end{figure*}

However, existing methods primarily cater to avatars with tight clothing and face challenges when extending to loose garments, as they only model body dynamics. Xiang \etal~\cite{xiang2021modeling, xiang2022dressing,xiang2023drivable} employ a dual-layer mesh representation to explicitly model garments.
While capable of rendering high-quality images, these approaches demand accurate geometry reconstructions, requiring dense cameras or RGB-D inputs.
Another family of research, e.g., \cite{zhao2022human, isik2023humanrf,jayasundara2023flexnerf}, focuses on capturing human performance with diverse garments using multi-view cameras but lacks the ability to generalize to unseen poses.

In contrast, our method relies solely on sparse multi-view videos as input and demonstrates the capacity to generalize garment dynamics under novel poses. The characteristics of our method and several representative existing methods are summarized in \reftab{tab:charac}.

\subsection{Garment Capture}
Capturing 3D garments is essential for various applications, including virtual try-ons and generative modeling.
Several methods have been developed to reconstruct static garments from a single image \cite{zhu2020deep, jiang2020bcnet, srivastava2022xcloth}. Recent works \cite{feng2022capturing, qiu2023rec} further enable us to capture dynamic garments from a monocular video. 
However, monocular videos inevitably introduce shape ambiguity for garment reconstruction and the temporal coherency of these methods can not be ensured.
Our research, however, focuses on recovering temporally coherent 3D garments for appearance learning from a sparse multi-view video, differing from these approaches.

Recent works from Xiang \etal~\cite{xiang2021modeling, xiang2022dressing, xiang2023drivable} rely on well-reconstructed garment geometries under dense views and register template garments to reconstructions using non-rigid ICP (Iterative Closest Point)~\cite{li2008global}.
\cite{li2021deep, habermann2020deepcap, habermann2023hdhumans} present approaches for predicting garment deformation from a single image, achieved by training a manually defined deformation graph to align with a pre-captured multi-view dataset. 
In contrast, we only have sparse views of short videos as input, and it is difficult to reconstruct the garment geometry accurately or pre-train a deformation-predicting network.

\subsection{Garment Animation}
\label{subsec:garment_animation}
Modeling realistic garment dynamics driven by non-rigid body motions is crucial for creating high-fidelity human avatars.
Research in this domain generally falls into two main categories: physics-based simulation \cite{baraff1998large, zeller2005cloth, vassilev2001fast, tang2013gpu, liu2017quasi} and data-driven methods \cite{de2010stable, lahner2018deepwrinkles, santesteban2019learning, wang2019learning, holden2019subspace, patel2020tailornet, vidaurre2020fully, bertiche2021deepsd, corona2021smplicit, zhang2021dynamic, li2022ncloth, pan2022predicting}.
Here, we focus on a recent study closely related to our work.
Instead of directly inferring animations at the mesh level, \cite{pan2022predicting} propose to transfer body motions to a set of bone transformations of a pre-computed garment skinning model.
However, their research primarily investigates the use of this skinning model for animating garment geometry for faster simulation.
In our paper, we delve deeper into the challenges of capturing and rendering garment dynamics in real-captured videos utilizing a similar model.
It is important to note that their approach is complementary to ours, as it provides a foundation for inferring garment bone motions for unseen body movements at test time.
\section{Method}

Our method takes as input a sparse multi-view video of a performer wearing a loose garment along with the corresponding garment template mesh.
Our goal is to construct an animatable loose-fitting avatar that supports photo-realistic rendering of not only body poses but also garment dynamics. 

\begin{figure*}[t!]
    \centering
    \includegraphics[width=\linewidth]{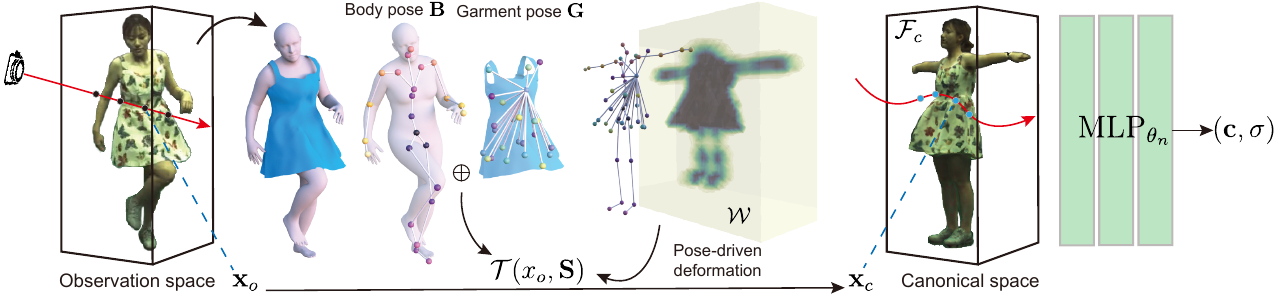}
    \caption{Overview of our rendering pipeline. For each sampled point in observation space, we first transform it back to the canonical space using a pose-driven deformation module conditioned on both body $\mathbf{B}$ and garment poses $\mathbf{G}$ and then query its color and density $(\mathbf{c}, \sigma)$ through a radiance field $\mathcal{F}_c$ defined in the canonical space.}
    \label{fig:rendering}
\end{figure*}

To achieve so, our method consists of three major steps. First, we build a garment rigging model from off-line physics-based simulation data, encoding possible garment shapes into low-dimensional garment poses (\refsec{subsec:garment_lbs}). 
Then, a novel garment fitting method is introduced for estimating the coarse garment dynamics from sparse multi-view videos $\mathcal{I}$ aided by such a rigging model. 
Please see \reffig{fig:gmlbs} for an overview of the above two steps.
Being equipped with both body and garment poses, we learn a generalizable pose-driven rendering network from $\mathcal{I}$, where both garment and body poses are exploited to drive a deformable neural radiance field (\refsec{subsec:nerf_rep} and \reffig{fig:rendering}). At test time, our method requires both body and garment poses to produce realistic rendering. We show how garment poses can be obtained for unseen body poses in \refsec{subsec:runtime_ani}.

\subsection{Garment Rigging Model}
\label{subsec:garment_lbs}
In this subsection, we first introduce the formulation of our garment rigging model and then show how to build it from simulation data. 
We adopt Linear Blend Skinning (LBS) for our garment rigging model.
In the LBS model, mesh deformations are driven by a set of bones. 
Given a template mesh $\mathcal{M}$ in rest pose with $\mathbf{v}_i$ denotes the position of its $i$-th vertex, the deformed $i$-th vertex $\mathbf{v}_{i}^{p}$ at configure $p$ can be represented as:
\begin{equation}
	\label{eq:lbs}
	\mathbf{v}_{i}^{p} = \sum_{j=1}^{B} w_{ij} \mathbf{T}_{j}^{p} \mathbf{v}_i
\end{equation}
where $\mathbf{T}_{j}^{p}$ represents $j$-th bone's transformation matrix at configure $p$, and $B$ is the number of bones. $\mathbf{W} = \{w_{ij}\}$ is the mesh skinning weights and each scalar $w_{ij}$ reflects the influence of $j$-th bone to the $i$-th vertex and satisfies $w_{ij}\geq 0$ and $\sum_{j=1}^{B}w_{ij}=1$.
At configure $p$, we refer $\mathbf{G}^{p} = \{\mathbf{T}^{p}_{j}\}_{j=1}^{B}$ as garment poses and denote the deformed mesh $\mathcal{M}^p=\text{LBS}(\mathcal{M}, \mathbf{W}, \mathbf{G}^{p})$ for compactness.


Given a set of deformed meshes $\{\mathcal{M}^p\}$, learning such an LBS model equals localizing $B$ bones in canonical space and solving the joint transformations $\mathbf{G}^{p}$, together with skinning weight matrix $\mathbf{W}$, that best explain the observations $\{\mathcal{M}^p\}$. To ensure the capacity of such LBS representation, we propose to learn it from diverse garment examples obtained from simulation.

\mparagraph{Physics-based Garment Simulation.} Specifically, we select several body motions with high dynamics covering walking, running, and dancing from the AMASS dataset~\cite{AMASS:2019}.
Together with the captured body motions from our multi-view video dataset, we then run the physics-based simulation to obtain diverse garment shapes $\{\mathcal{M}^p\}$, as shown in \reffig{fig:gmlbs}.

\mparagraph{Skinning Decomposition.} Given the simulated mesh sequences, we further encode the garment dynamics from high-dimensional vertex space into an LBS model using the state-of-the-art skinning decomposition method Smooth Skinning Decomposition with Rigid Bones (SSDR)~\cite{le2012smooth}.
Specifically, SSDR solves bone transformations $\mathbf{G}^{p}$ and skinning weights $\mathbf{W}$ by minimizing the mesh reconstruction loss.
We refer the readers to \cite{le2012smooth} for more details about decomposition.
Following~\cite{pan2022predicting}, we refer to the extracted skeleton as virtual bones since these joints may not actually located at mesh surfaces.
Such an LBS model forms a compact representation of the underlying geometry, enabling us to capture and animate garment dynamics via garment poses rather than high-dimensional vertices.

\subsection{Garment Dynamics Fitting}
\label{subsec:garment_fitting}
Being equipped with the rigging model, we are now able to estimate garment dynamics from sparse multi-view videos using bone transformations.
Formally,  we estimate garment poses $\mathbf{G}^{t}$ for each timestep $t$, while ensuring the alignments between garment meshes and sparse view observations $\mathcal{I}$. 

However, such a task still poses significant challenges due to the inherent complexities of fabric behavior.  
Garment deformations, which are complex and non-linear, construct a large space of valid shapes.
Capturing folds and drapes of garments, which constantly change with body movements, is ill-posed. 
Unlike body mesh fitting where explicit keypoints detection and kinematic constraints are often used, limited regularizations can be used for estimating garment dynamics. 
Moreover, ensuring temporal consistency across frames during estimation is also challenging~\cite{qiu2023rec}.

To address the above challenges, we develop a novel optimization method based on differentiable rendering~\cite{liu2019softras, ravi2020pytorch3d} where we align 3D garment shapes to a set of robust 2D cues, i.e., garment silhouettes, image normals, and optical flows.
We show each of these cues boosts fitting performance at one specific aspect and is essential for obtaining valid garment dynamics (see \refsec{subsec:ablation}). 

To begin with, we optimize the difference between rendered garment masks and detected 2D silhouettes. 
However, silhouette loss can only offer a matched 3D outer shape, leading to invalid folds and drapes.
To predict more accurate folds, we further incorporate normal cues predicted from 2D images.
To enhance the temporal consistency across frames, mesh deformations are projected in 2D space to match pixel movements of optical flow.
The right part of \reffig{fig:gmlbs} illustrates our fitting step.

Overall, the optimization problem can be formulated as:
\begin{equation}
	\label{eq:losses}
	\begin{aligned}
		&\underset{\mathbf{G}^t}{\text{minimize}} \mathcal{L}_{fit}(\text{LBS}(\mathcal{M}, \mathbf{W}, \mathbf{G}^t)),\\
		&\mathcal{L}_{fit} = \lambda_m \mathcal{L}_M + \lambda_n \mathcal{L}_N + \lambda_{of} \mathcal{L}_{of}.
	\end{aligned}
\end{equation}
\noindent where $\mathcal{L}$ are different terms of loss terms and $\lambda$ represent their weights. We illustrate each loss term individually in the following.

\mparagraph{Garment Silhouette Loss.}
We use~\cite{li2020self} to get garment silhouettes $I^M$ and silhouette loss in multi-views is defined as:
\begin{equation}
	\mathcal{L}_M = \sum_v^{N^v}||\mathcal{DR}(\text{LBS}(\mathcal{M}, \mathbf{W}, \mathbf{G}^t)), c_v) - I^M_v||^2_2,
\end{equation}

\noindent where $\mathcal{DR}$ is the differentiable renderer~\cite{liu2019softras}, $N^v$ is the number of views, and $c_v$ is the camera parameters of the $v$-th view. 

\mparagraph{Image Normal Loss.}
We use normal predictor proposed by~\cite{xiu2022icon} to estimate the normal map $I^N$ from 2D images and formulate the normal projection loss as:
\begin{equation}
    \mathcal{L}_N = \sum_v^{N^v}||\mathcal{N}^p(\text{LBS}(\mathcal{M}, \mathbf{W}, \mathbf{G}^t)), c_v) - I^N_v*I^M_v||^2_2,
\end{equation}
where $\mathcal{N}^p(\cdot, c)$ is the rendered normals of the visible mesh faces in view $c$. We also multiple the garment mask $I^M$ on the predicted normal map. 

\mparagraph{Optical Flow loss.}
Once we get the fitted geometry $\mathcal{M}^{t-1}$ of ($t$-1)-th frame, we can use it as the initialization for the next timestep and leverage the predicted 2D optical flow $I^{of}$ as temporal constraint between frames.
We employ the commonly-used RAFT~\cite{teed2020raft} to estimate the optical flow between two adjacent frames, and the optical flow loss is defined as:
\begin{equation}
    \begin{aligned}
    \mathcal{L}_{of} = &\sum_v^{N^v}||\mathcal{C}^p(\text{LBS}(\mathcal{M}, \mathbf{W}, \mathbf{G}^t), c_v) \\
    &\quad\quad- (\mathcal{C}^p(\mathcal{M}^{t-1}, c_v)+\delta_v)||_2^2, \\
    &\delta_v = I^{of}_v(\mathcal{C}^p(\mathcal{M}^{t-1}, c_v)),
    \end{aligned}
\end{equation}
where $\mathcal{C}^p(\cdot, c)$ means visible projected coordinates in the screen space in the camera view $c$. We optimize the projected mesh movement between $\mathcal{M}^{t-1}$ and the newly predicted shape to match the optical flow value $\delta$ in $I^{of}$. Notably, for the first frame, we only use $\mathcal{L}_M$ and $\mathcal{L}_N$ because there is no previous frame for estimating optical flow.

\subsection{Pose-driven Rendering Network}
\label{subsec:nerf_rep}
Given a sparse multi-view video $\mathcal{I}$, along with the estimated body and garment poses, we aim to build a generalizable NeRF-based rendering module with controllability over both body and garment dynamics.

Following previous works \cite{peng2021animatable}, we represent a moving person with a neural radiance field $\mathcal{F}_c$ in the canonical space.
A deformation module $\mathcal{T}$ is used to map points from observation space back to the canonical space for efficient rendering. The challenge here is how one learn such deformation accurately. Previous template-based methods \cite{peng2021neural, peng2021animatable} drives near-surface points using mesh deformations. However, simply applying such strategy for modeling both body and garment movements leads to sub-optimal performance as it requires us to not only estimate accurate geometries for both parts but also resolve the body-garment mesh penetrations. 

To address the above challenges, we propose to learn a pose-driven deformation field conditioned on both 3D body and garment poses.  Specifically, the color and density of point $\mathbf{x}_{o}$ in the observation space is defined as:
\begin{equation}
    (\mathbf{c}(\mathbf{x}_o), \sigma(\mathbf{x}_o)) = \mathcal{F}_c(\mathcal{T}(\mathbf{x}_o, \mathbf{G}, \mathbf{B})),
\end{equation}
\noindent where $\mathcal{T}$ is a backward deformation module that takes both body pose $\mathbf{B}$ and garment pose $\mathbf{G}$ as input, and outputs the point $\mathbf{x}_{c}$ in canonical space.
$\mathcal{F}_c$ is a 
multi-layer perceptrons $\text{MLP}_{\theta_n}$ that takes point in the canonical space $\mathbf{x}_{c}$ as input and outputs its color value $\mathbf{c}$ and density $\sigma$.

\mparagraph{Pose-driven Deformation Module.}  Instead of modeling body and garment motions independently, which requires us to further infer their interactions, we concatenate them together to form a unified pose $\mathbf{S}$, as shown in \reffig{fig:rendering}. 
Then, the transformation $\mathcal{T}$ is defined as an inverse LBS function that deforms points in the observation space to canonical space:
\begin{equation}
	\mathcal{T}(\mathbf{x}_o, \mathbf{S}) = \sum_k^{J}w_o^{k} \mathbf{T}_{k}^{-1} \mathbf{x}_o,
\end{equation}
where $\mathbf{T}_{k}^{-1}$ is the inverse transformation matrix of $k$-th joint of $\mathbf{S}$, $J$ is the number of joints and $w_o^{k}$ is inverse blend weights of $\mathbf{x}_o$ respects to $k$-th joint. 
However, this formulation requires us to solve each observation space an inverse blend weight field, which is computation intensive. Moreover, the learned model may be over-fitted to training frames and generalizes poorly to unseen poses. 

To address the above challenge, we follow~\cite{weng2022humannerf} to define the inverse blend weight field in one shared canonical space. Specifically, we use a $\text{CNN}_{\theta_t}$ parameterized by ${\theta_t}$ to generate the inverse blend weight field $\mathcal{W}$ from a fixed random latent code. Then $w_o^k$ is computed as:
\begin{equation}
	w_o^k =\frac{w_c^k\mathbf{T}_i^{-1} \mathbf{x}_o}{\sum_{i=1}^{J} w_c^i \mathbf{T}_i^{-1} \mathbf{x}_o},
\end{equation}
where $w_c^k =\mathcal{W}(\mathbf{T}_i^{-1} \mathbf{x}_o)$ denotes the queried value in the blend weight field $\mathcal{W}$ at the position $\mathbf{T}_i^{-1} \mathbf{x}_o$ in the canonical space.

\mparagraph{NeRF Optimization.}
\label{subsec:nerf_optim}
Based on the above NeRF representation, we are able to optimize the canonical appearance module $\mathcal{F}_c$, together with the inverse blend weights volume $\mathcal{W}$ using volume rendering techniques~\cite{mildenhall2020nerf}. 
Specifically, we cast a ray $\mathbf{r}$ at observation space and minimize the mean squared error (MSE) between the rendered RGB color $\tilde{\mathbf{C}}$ with the ground truth $\mathbf{C}$:
\begin{equation}
	\mathcal{L}_{\text{MSE}} = \sum_{i\in N^f}\sum_{j\in N^v}\sum_{r\in \mathcal{R}}||\tilde{\mathbf{C}}_{i,j}(\mathbf{r}) - \mathbf{C}_{i,j}(\mathbf{r})||_2^2,
\end{equation}
where $N^f, N^v$ are numbers of frames and views, and $\mathcal{R}$ is the set of rays passing through images pixels. We also adopt a perceptual loss LPIPS~\cite{zhang2018unreasonable} and the final loss of our rendering network is $\mathcal{L}=\mathcal{L}_{\text{MSE}}+\lambda\mathcal{L}_{\text{LPIPS}}$.

\subsection{Test Time Animation}
\label{subsec:runtime_ani}
At test time, our method requires both novel body and garment poses to produce photo-realistic rendering.
Here, we show several ways to obtain valid garment poses. First, novel garment poses can be estimated from a newly-come multi-view video, as shown in \reffig{fig:npg_comp}. In such a situation, garment poses serve as control signals to reproduce the garment dynamics presented in the newly-come multi-video under arbitrary viewpoints without retraining the rendering model.
Second, garment poses can be borrowed from our simulated and decomposed garment sequences to drive unseen dynamics, shown in \refsec{subsec:novel_pose_sim}.
Thirdly, we can run the physics-based simulation to obtain new garment shapes, and the garment poses can be further obtained via mesh fitting.
Moreover, as discussed in \refsec{subsec:garment_animation}, existing work~\cite{pan2022predicting} has already explored the possibility of inferring garment poses for unseen body poses using neural networks and achieved impressive results.
Their work motivates us and serves as a strong complementary as they mainly focus on animating garment poses for novel body motion and our method creates realistic garment rendering models from real data.
\section{Dataset}

The commonly used multi-view video datasets for human modeling, such as H3.6M~\cite{h36m_pami} and ZJU-MoCap~\cite{peng2021neural}, focus on subjects in tight clothing and are thus unsuitable for evaluating our method.
We create a new dataset comprising 12 dynamic videos featuring a performer in 5 different loose clothing such as dresses and skirts.
We captured at least 14 camera views for every motion, selecting eight uniformly distributed cameras for training and the rest for testing novel view performance.
Each garment is represented in at least two sequences showcasing varied body movements to facilitate evaluation in novel poses.
When we exploit one of the videos for training, the other video will be used as test data for novel pose evaluation.
The length of each sequence ranges from 400 to 800 frames.
For each sequence, we generate binary masks for the human body and garment using~\cite{li2020self} and estimate 3D body poses with~\cite{easymocap}.
The mesh template for each garment is derived from static reconstruction, and more details can be found in our supplementary material.
\section{Experiments}

\begin{figure}[t!]
    \centering
    \includegraphics[width=\linewidth]{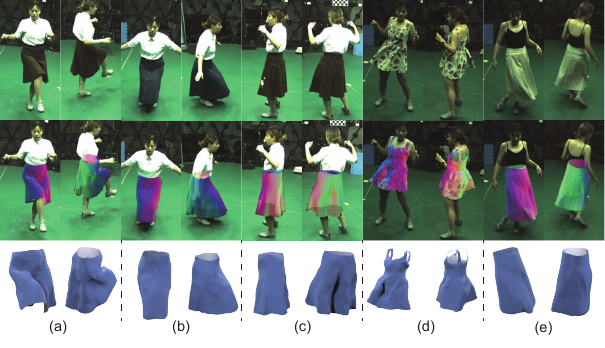}
    \caption{Visualizations of our garment fitting results, where the 1$^{\text{st}}$ and the 3$^{\text{rd}}$ rows show the inputs and rendered garment geometry, respectively. We also overlay the rendered mesh geometry onto input images to illustrate the alignments (the 2$^{\text{nd}}$ row).}
    \label{fig:fit_res}
\end{figure}

\begin{figure*}[ht]
    \centering
    \includegraphics[width=\linewidth]{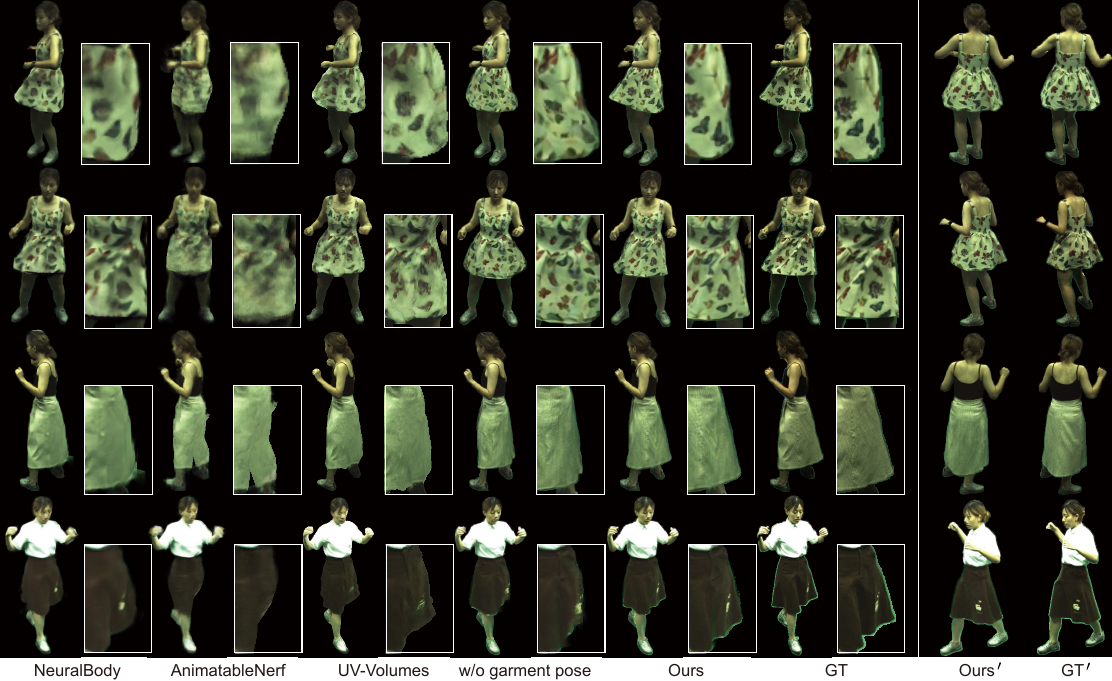}
    \caption{Visual comparisons on novel view synthesis. We also show an extra view of our results denoted as Ours$'$ and the GT$'$.}
    \label{fig:nvs_comp}
\end{figure*}

\begin{figure*}[ht]
 \centering
 \includegraphics[width=\linewidth]{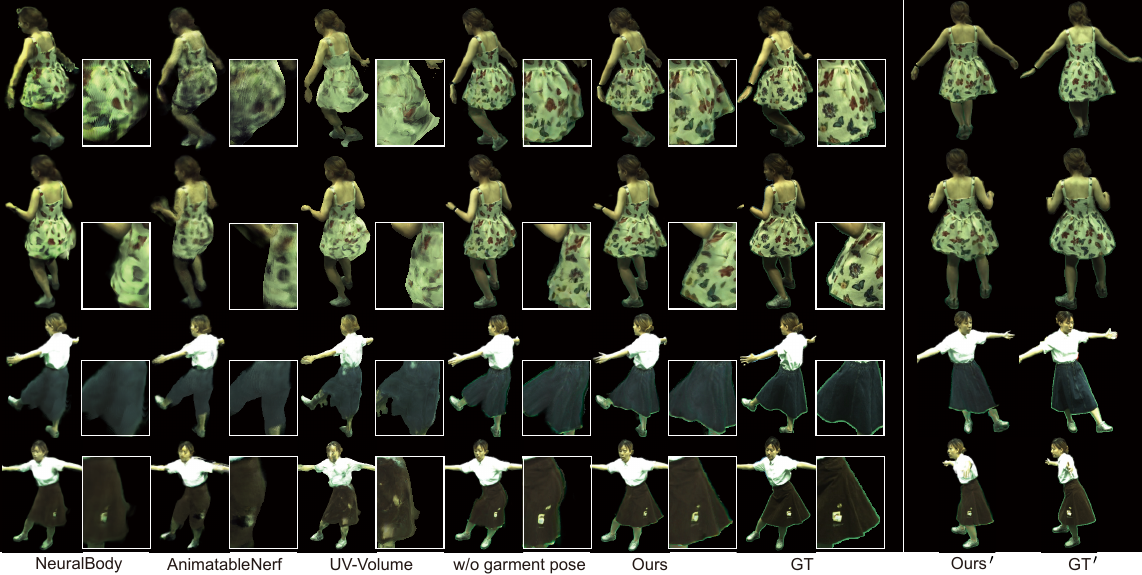}
 \caption{Visual comparisons on novel pose generation. We also show an extra view of our results denoted as Ours$'$ and the GT$'$.}
 \label{fig:npg_comp}
\end{figure*}

\subsection{Results of Garment Fitting}
In \reffig{fig:fit_res}, a collection of garment fitting results is presented, demonstrating the efficacy of the novel fitting method proposed in \refsec{subsec:garment_fitting}. Though our method fails to capture details such as wrinkles, the fitted garment meshes align with the input images in silhouettes and exhibit coarse folds similar to observations (\reffig{fig:fit_res} (a) and (c)).
Our method also successfully captures dynamics such as skirt swings (\reffig{fig:fit_res} (b) (d) and (e)). The recovered dynamics is temporally coherent and we recommend the readers to see our supplementary video for more visualization.


\subsection{Image Quality Evaluation}

\mparagraph{Baselines:}
To validate our method, we compare it against several state-of-the-art human modeling methods: NeuralBody~\cite{peng2021neural}, Animatable-NeRF~\cite{peng2021animatable} and UV-Volumes~\cite{chen2023uv}.
NeuralBody~\cite{peng2021neural} anchors feature vectors on a deformable mesh template and generates the radiance field using forward skinning.
Animatable-NeRF \cite{peng2021animatable} learn a backward neural blend weights field conditioned on 3D body poses and points are transformed back to canonical space for rendering.
Different from previous methods~\cite{peng2021neural,peng2021animatable} that predict pixel colors directly, UV-Volumes~\cite{chen2023uv} render UV coordinates of each pixel and the RGB values are then queried from a neural texture stack.
We also compare our method with a variant that does not use our proposed garment rigging model, i.e., using body poses $\mathbf{B}$ only in the pose-driven rendering network, which is similar to a multi-view variant of ~\cite{weng2022humannerf}.

\mparagraph{Metrics:}
In line with~\cite{chen2023uv}, our evaluation employs the Peak Signal-to-Noise Ratio (PSNR) for pixel-level measurement and Learned Perceptual Image Patch Similarity (LPIPS) for patch feature-level assessment.
Additionally, we utilize Intersection over Union (IoU) to assess the quality of the generated avatar's outer shape.
Furthermore, we utilize Deep Image Structure and Texture Similarity (DISTS)~\cite{ding2020image} as a metric for evaluating texture similarity.

\begin{table}[t]
    \caption{Quantitative comparisons on novel view synthesis. We color cells that have the \colorbox{red!25}{best} and \colorbox{orange!25}{second best} scores. $\mathrm{LPIPS*}=\mathrm{LPIPS}\times10, \mathrm{DISTS*}=\mathrm{DISTS}\times10$.}
    \centering
    \scalebox{0.85}{
    \begin{tabular}{c|c|c|c|c}
    \hline
                                       & \multicolumn{1}{l|}{$\mathrm{PSNR}\uparrow$} & \multicolumn{1}{l|}{$\mathrm{IoU}\uparrow$} & \multicolumn{1}{l|}{$\mathrm{LPIPS*}\downarrow$} & \multicolumn{1}{l}{$\mathrm{DISTS*}\downarrow$} \\ \hline
    NeuralBody~\cite{peng2021neural}   & \cellcolor{red!25}28.45 & 78.95 & 0.4239 & 1.1132 \\ \hline
    Anim-Nerf~\cite{peng2021animatable}& 26.67 & 72.13 & 0.5866 & 1.5989 \\ \hline
    UV-Volumes~\cite{chen2023uv}       & 26.82 & \cellcolor{red!25}88.81 & 0.3231 & 1.0078 \\ \hline
    Ours w/o $\mathbf{G}$              & 27.12 & 85.27 & \cellcolor{orange!25}0.3026 & \cellcolor{orange!25}0.9208 \\ \hline
    \textbf{Ours}                      & \cellcolor{orange!25}27.40 & \cellcolor{orange!25}86.14 & \cellcolor{red!25}0.2846 & \cellcolor{red!25}0.8838   \\ \hline
    \end{tabular}}
    \label{tab:nvs}
\end{table}

\begin{table}[t!]
    \caption{Quantitative comparisons on novel pose synthesis. We color cells that have the \colorbox{red!25}{best} and \colorbox{orange!25}{second best} scores. $\mathrm{LPIPS*}=\mathrm{LPIPS}\times10, \mathrm{DISTS*}=\mathrm{DISTS}\times10$.}
    \centering
    \scalebox{0.85}{
    \begin{tabular}{c|c|c|c|c}
    \hline
                                       & \multicolumn{1}{l|}{$\mathrm{PSNR}\uparrow$} & \multicolumn{1}{l|}{$\mathrm{IoU}\uparrow$} & \multicolumn{1}{l|}{$\mathrm{LPIPS*}\downarrow$} & \multicolumn{1}{l}{$\mathrm{DISTS*}\downarrow$} \\ \hline
    NeuralBody~\cite{peng2021neural}   & \cellcolor{orange!25}25.31 & 75.97 & 0.5486 & 1.4317 \\ \hline
    Anim-Nerf~\cite{peng2021animatable}& 24.70 & 71.19 & 0.6458 & 1.7347 \\ \hline
    UV-Volumes~\cite{chen2023uv}       & 24.39 & \cellcolor{red!25}82.53 & 0.4899 & 1.4309 \\ \hline
    Ours w/o $\mathbf{G}$              & 24.97 & 80.95 & \cellcolor{orange!25}0.4095 & \cellcolor{orange!25}1.1275 \\ \hline
    \textbf{Ours}                      & \cellcolor{red!25}25.45 & \cellcolor{orange!25}82.23 & \cellcolor{red!25}0.3774 & \cellcolor{red!25}1.0744   \\ \hline
    \end{tabular}}
    \label{tab:npg}
\end{table}

\mparagraph{Novel View Synthesis:}
\reffig{fig:nvs_comp} shows the visual comparisons with baselines~\cite{peng2021neural,peng2021animatable,chen2023uv} and our variant without garment pose $\mathbf{G}$.
Baselines struggle to produce clear textures and accurate shapes for dynamic dresses. 
Benefiting from the estimated garment pose, our method successfully maintains consistency in both garment appearances and geometries across views.
Quantitative comparisons are detailed in \reftab{tab:nvs}. LPIPS scores indicate the superiority of our method in generating textures that agree with human perceptions and DISTS scores show our ability to generate similar textures with the the ground truth.  
Furthermore, IoU scores demonstrate that our method achieves significantly more accurate outer shapes compared to~\cite{peng2021neural}.
Note that, UV-Volume~\cite{chen2023uv} additionally uses a silhouette loss for training, thus generating accurate outer shapes. However, their textures do not match the quality of ours.

\mparagraph{Novel Pose Generation:}
We estimate body and garment poses from the test motion sequences and drive trained avatars to generate novel poses for our method.
Visual comparisons are shown in \reffig{fig:npg_comp} and quantitative results are presented in \reftab{tab:npg}.
Existing methods like NeuralBody and UV-Volumes show limited generalizability with loose garments, while our method exhibits superior visual quality, featuring clear textures and correct shapes.
It is noteworthy that, without the proposed garment rigging model, the generated dresses will have pants-like shapes, as shown in 3$^\text{rd}$ row \reffig{fig:npg_comp}.
The superiority shown quantitatively further demonstrates the generalizability of our method.

\begin{figure}[t!]
    \centering
    \includegraphics[width=\linewidth]{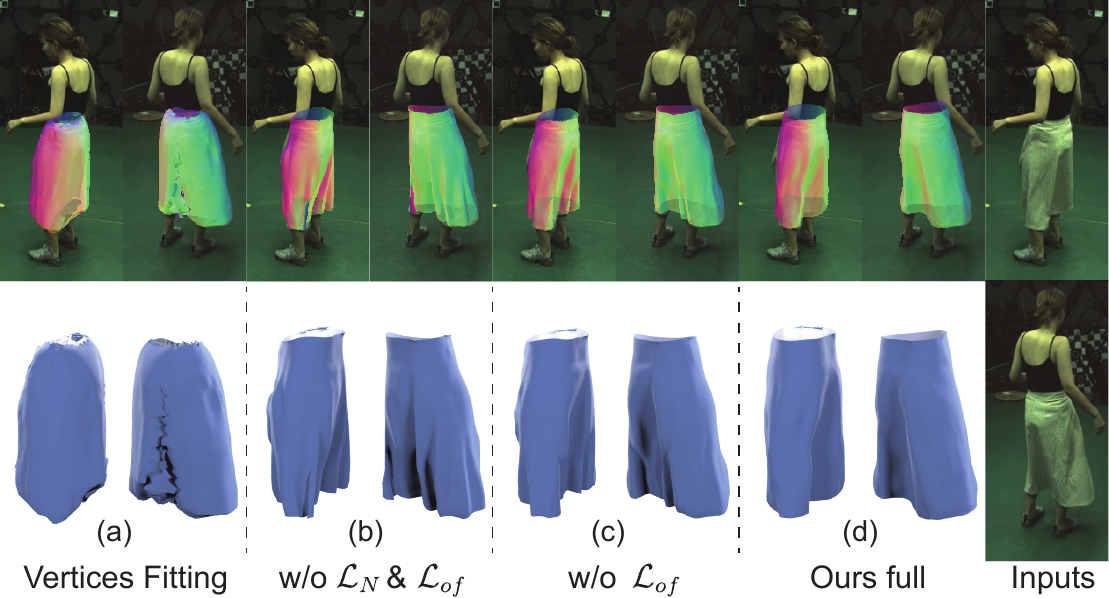}
    \caption{Visual comparisons of fitted mesh geometries. We compare our full method with (a) directly fitting mesh vertices, (b) without using normal $\mathcal{L}_N$ and optical flow $\mathcal{L}_{of}$ losses, and (c) without using the optical flow loss.}
    \label{fig:fit_ablation}
\end{figure}

\subsection{Ablation Study}
\label{subsec:ablation}
We provide ablation studies for the proposed garment fitting step to validate the robustness of our method.
As shown in \reffig{fig:fit_ablation}, comparisons between our full model and its ablated versions (a)-(c) highlight the efficacy of the introduced components.
Ablation (a) abandons our garment rigging model and directly uses garment mesh vertices for fitting, resulting in invalid mesh geometries with serve self-penetrations.
In ablation (b), both normal loss and optical flow loss are disabled.
While the outer shape of the garment mesh aligns with the 2D silhouettes, the dress folds do not match image observations. 
In ablation (c), we remove optical flow loss.
Using normal loss only, the optimization for the geometry shape simply deforms the vertices to meet the normal prediction, resulting in garment self-penetrations.
In contrast, our full model recovers the desired garment dynamics under the constraints of predicted optical flow.
The incorporation of optical flow loss $\mathcal{L}_{of}$ enhances the temporal consistency of the fitted geometry sequence and we show this improvement in the supplementary video.

\begin{figure}[t!]
    \centering
    \includegraphics[width=\linewidth]{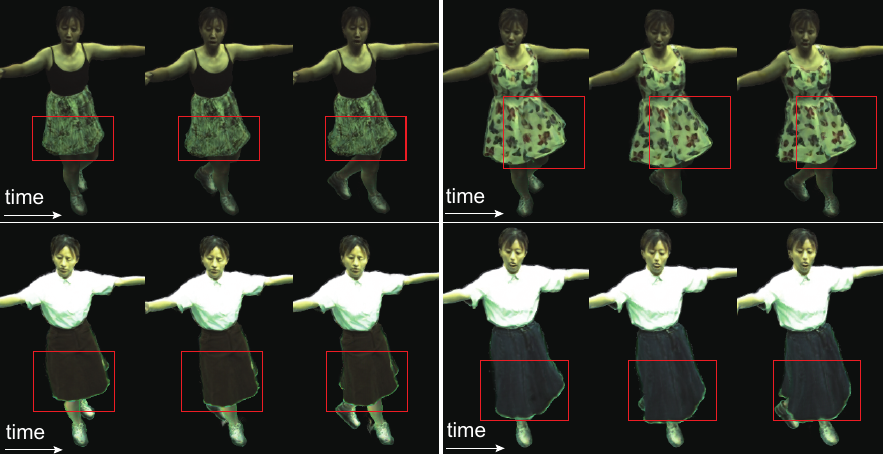}
    \vspace{-0.3cm}
    \caption{The generated renderings under novel pose sequences with garment poses driven by physics-based simulations. Please see the highlights for the dynamic differences.}
    \vspace{-0.3cm}
    \label{fig:np_sim}
\end{figure}

\subsection{Novel Pose Generation Using Simulation Data}
\label{subsec:novel_pose_sim}
To further show the generalization ability of our rendering method, we generate images driven by garment poses produced by the physics-based simulation.
Specifically, we select one certain dance body motion as well as four dress geometries for the simulation and compute the garment pose for each dress.
Using the garment poses together with the body pose, we drive four pre-trained avatars to generate novel pose sequences.
As shown in \reffig{fig:np_sim}, our method can generate plausible dynamics for loose garments for unseen motions.
\section{Discussion}

\mparagraph{The influence of physics-based simulation parameters.} One of the most important steps in the proposed method is to run a stable and valid physics-based simulation to constitute a garment dataset.
However, in-valid simulation parameters may lead to simulation failures.
To address this problem, we conduct the simulation on each garment using Houdini Vellum~\cite{houdini} with default settings except for the fabric's bending stiffness and the simulator's time scale since these two parameters influence the simulated results the most.
The time scale is adjusted based on their respective motion capture FPS. While the bending stiffness controls how soft the material is. We choose the bending stiffness to be 1e-4 for all garments since it corresponds to a soft material. This is based on the observations that using soft materials in simulation can make sure that the garment rigging model is able to express stiff status in capturing. However, if we start with a stiff material in simulation (i.e., larger bend stiffness), the garment rigging model may not be able to capture soft garment dynamics.  

\mparagraph{Conclusion.}
We introduced \anidress, an innovative approach for building expressive loose-dressed animatable avatars, capable of synthesizing images in novel views and novel poses.
By leveraging a garment rigging model, our method captures and renders garment dynamics using a set of bone transformations.
Technically, we introduce a novel method for estimating garment poses to sparse multi-view videos, aided by 2D cues.
To provide controllability over both body and garment dynamics, a pose-driven neural radiance field conditioned on both body and garment parts is introduced to render high-quality images.
We build a new dataset consisting of loose-dressed performers in diverse body motions and demonstrate the superior performance of our method over baselines via extensive experiments.

\mparagraph{Limitations and Future Works.}
Our method requires a template mesh for building the rigging model. So far, obtaining such template mesh requires manual correction.
In the future, we will explore the possibility of automatic template mesh reconstruction.
Moreover, our method relies on the garment rigging model built from simulation data. So far, the rigging model needs to be rebuilt for different garment types. In the future, we will explore the possibility of building a generalizable rigging model that can be applied to different garment types.
Besides, we build our rendering module upon volume rendering where points far away from the garment surface may also contribute to the rendering.
This may cause ghosting effects in challenging novel pose scenarios. Geometry constraints can be used to further regularize the NeRF optimization process.
{
    \small
    \bibliographystyle{ieeenat_fullname}
    \bibliography{main}
}
\clearpage
\setcounter{page}{1}
\twocolumn[{%
\renewcommand\twocolumn[1][]{#1}%
\maketitlesupplementary
\begin{center}
    \centering
    \includegraphics[width=\linewidth]{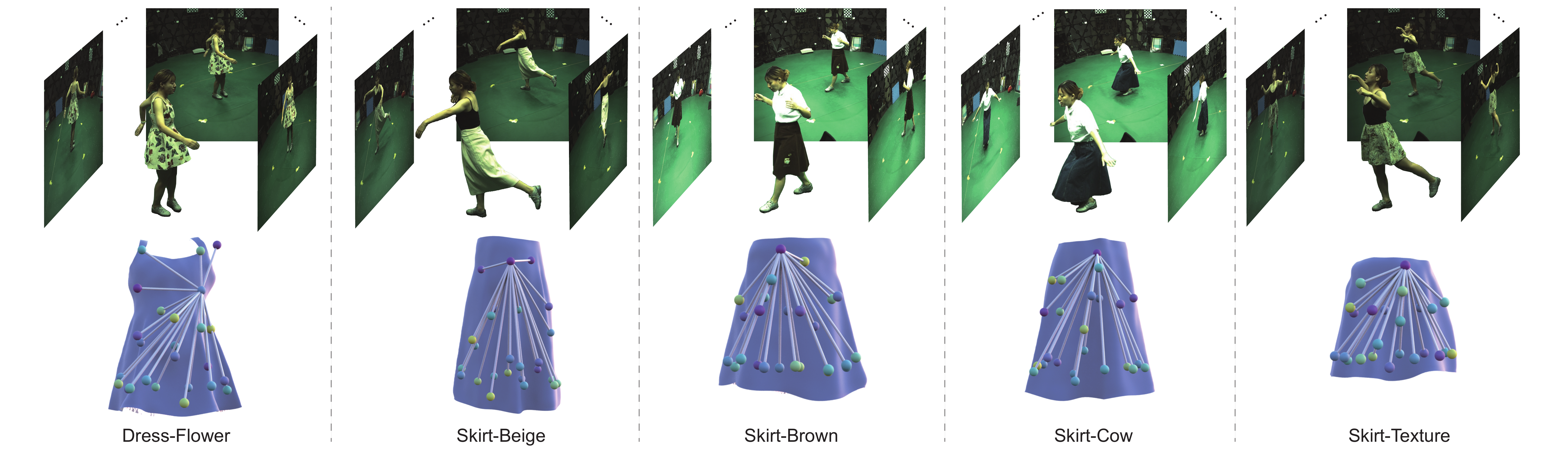}
    \captionof{figure}{Overview of our captured dataset. We show the captured multi-view images for each garment in row 1 and their corresponding rigging model in row 2.
    Specifically, we captured 5 different loose clothes denoted as ``Dress-Flower'', ``Skirt-Beige'', ``Skirt-Brown'', ``Skirt-Cow'' and ``Skirt-Texture'' respectively. }
    \label{fig:dataset}
\end{center}
}]

\section{Dataset Overview}
\label{sec:dataset}

\reffig{fig:dataset} presents an overview of our captured dataset, encompassing 5 different loose clothes, namely ``Dress-Flower'', ``Skirt-Beige'', ``Skirt-Brown'', ``Skirt-Cow'' and ``Skirt-Texture''.
For each garment, we recorded a minimum of two sequences, referred to as ``Seq-1'' and ``Seq-2''. Additionally, \reffig{fig:dataset} shows the garment template mesh and its associated rigging model in the second row.

\section{Implementation Details}
\label{sec:imp_de}
We provide more implementation details in this section.

\mparagraph{Garment T-pose Reconstruction.}
Our method necessitates a template mesh for each garment to run physics-based simulations.
Contrary to previous approaches \cite{habermann2021real} which relied on costly 3D scanners to obtain such templates, we adopt a more economical approach utilizing recent progress in neural surface reconstruction.
We employ a monocular camera to photograph subjects in T-pose and further calibrate the cameras with structure-from-motion techniques~\cite{schoenberger2016mvs, schoenberger2016sfm}.
The signed distance function in T-pose is reconstructed using NeuS~\cite{wang2021neus} and the mesh is extracted via marching cubes~\cite{lorensen1998marching}.
To isolate garment parts, we initially segment the full-body mesh using coarse labels from 2D human parsing ~\cite{li2020self}, followed by manual corrections for inaccuracies.
The reconstructed mesh is then downsampled to a vertex count between 6K and 9K.

\mparagraph{Garment Rigging Model Extraction.}
To build a rigging model, we first accumulate a variety of body movements for physics simulations.
We utilize 6,000 frames of high-quality, self-penetration-free human motion data from AMASS~\cite{AMASS:2019}, including walking, running, and dancing.
This is supplemented with 4,000 frames of our own captured body motion.
We conduct the simulation on each garment using Houdini Vellum~\cite{houdini} with default settings except for the fabric's bending stiffness and the simulator's time scale.
The bending stiffness is set to 1e-4 for all garments, offering a balance between dynamic richness and avoiding unnatural simulations.
The time scale is adjusted to $1.5$ for AMASS motions and $0.3$ for our captures, based on their respective motion capture FPS.
After simulation, we extract a garment rigging model using the open-source tool provided by~\cite{le2012smooth} from the simulated garment mesh sequence.
Specifically, we set the iteration count to 50 to ensure convergence. The number of bones $B$ to 25 for each garment empirically.

\begin{figure*}[t!]
    \centering
    \includegraphics[width=\linewidth]{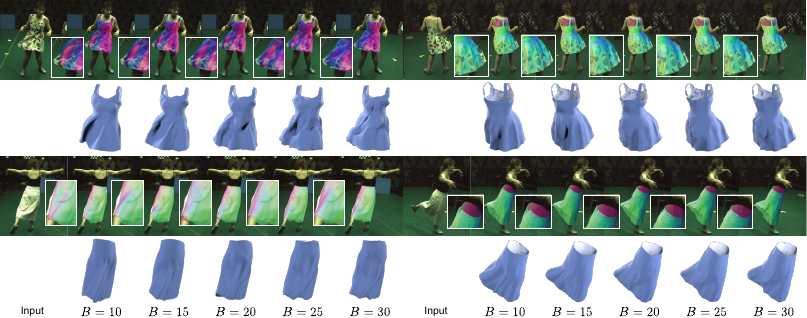}
    \caption{Garment fitting results with different numbers of garment bones $B$. In the 1$^{\text{st}}$ row, we show the original image and the rendered normal images of captured meshes.
    The fitted meshes are shown in the 2$^{\text{nd}}$ row for each case.}
    \label{fig:numj}
\end{figure*}

\mparagraph{Garment Dynamics Fitting.}
For the first frame, we only use silhouette loss and normal loss since there is no optical flow prediction for the first frame.
To estimate the garment poses at $t$-th frame, we use the garment poses from ($t$-1)-th frame for initialization and optimize it using all three losses. Specifically, we set loss weights as $\lambda_M=2.0, \lambda_N=0.1, \lambda_{of}=3.0$.

\mparagraph{Pose-driven Rendering Network.}
For the total training loss function, we set the weight $\lambda=0.2$ for all sequences.
Except for the appearance neural radiance filed $\mathcal{F}_c$ in canonical space, our pose-driven rendering network needs to jointly optimize a motion deformation filed $\mathcal{W}$ conditioned on both body and garment pose.
To better solve the motion field, following~\cite{weng2022humannerf}, we initialize $\mathcal{W}$ with the forward skinning weights given by the canonical T-pose and ask the network to learn the residue.
For the body part, we also use the combination of ellipsoidal Gaussian around each body bone, like~\cite{weng2022humannerf}.
However, for our garment rigging model, the inside `virtual' bones are not tied to the garment surface (as shown in \reffig{fig:dataset}).
Therefore, we initialize $\mathcal{W}$ with the skinning weights around garment T-pose mesh vertices for the garment part.


\section{More Experiment Results}
\subsection{Garment Dynamics Fitting}
\mparagraph{Number of Bones.} 
The number of garment bones $B$ serves as a critical hyperparameter in constructing rigging models. For all experiments presented in our main paper, we set $B$ as 25.
In this subsection, we evaluate the fitting performance of our method on ``Dress-Flower-Seq3'' and ``Long-Beige-Seq1'' using different numbers of garment bones (10, 15, 20, 25, 30).
The fitting results are presented in \reffig{fig:numj}.
We only provide qualitative results, as there is no ground-truth 3D garment geometry.
The first row displays the original images and the corresponding normal images of the fitted meshes, while the second row shows the reconstructed meshes.
As indicated in \reffig{fig:numj}, models with fewer bones lack the expressiveness to capture garment folds and shapes accurately.
Generally, models with more joints yield more precise fitting results.
Our approach primarily recovers folds and shapes observable in the images, excluding the finer details of clothing wrinkles.
This limitation arises mainly due to two factors.
First, a more expressive rigging model capable of detailing wrinkles is required, which is beyond the scope of our current method.
The LBS model, derived from skinning decomposition, represents the folds and shapes of simulated data but omits local wrinkles, making it less effective for capturing local deformations.
Second, our method depends on 2D cues like image normals and optical flows predicted from \cite{xiu2022icon, xiu2023econ, teed2020raft} for capturing temporal consistent 3D folds. 
While these cues are reliable for large folds and shapes, they are less accurate in depicting local wrinkles.

\begin{figure}[t!]
    \centering
    \includegraphics[width=\linewidth]{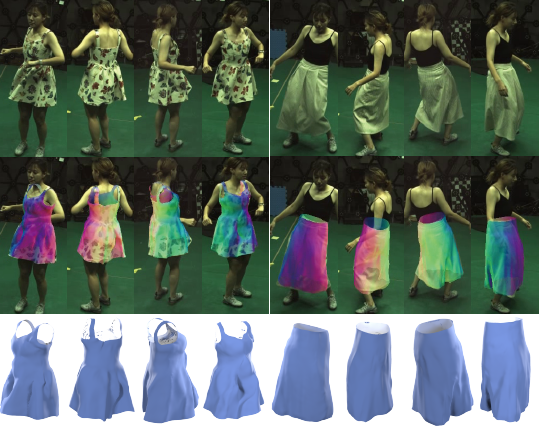}
    \caption{Garment fitting results under 4 cameras. We show the original image in the 1$^{\text{st}}$ row and the rendered normal images of captured meshes in the 2$^{\text{nd}}$ row. The fitted meshes are shown in the 3$^{\text{rd}}$row.}
    \label{fig:numcams}
\end{figure}

\begin{table*}[t!]
\caption{Quantitative results of novel view synthesis. Generally, our method reports slightly lower PSNRs than NB (NeuralBody) \cite{peng2021neural} while outperforming all baselines on LPIPS \cite{zhang2018unreasonable} and DISTS \cite{ding2020image}.
We color cells that have the \colorbox{red!25}{best} and \colorbox{orange!25}{second best} scores.
(AN: Anim-Nerf~\cite{peng2021animatable}, NB: NeuralBody~\cite{peng2021neural}, Uv: UV-NeRF~\cite{chen2023uv} and w/o $\mathbf{G}$: without using garment pose.)}
\resizebox{\textwidth}{!}{%
\begin{tabular}{c|c|lllll|lllll|lllll}
    \hline
    \multirow{2}{*}{Garment} &
    \multirow{2}{*}{Sequence} &
    \multicolumn{5}{c|}{$\mathrm{PSNR}\uparrow$} &
    \multicolumn{5}{c|}{$\mathrm{LPIPS}\downarrow$ \cite{zhang2018unreasonable}} &
    \multicolumn{5}{c}{$\mathrm{DISTS}\downarrow$~\cite{ding2020image}} \\
    & &
    \multicolumn{1}{c}{AN} &
    \multicolumn{1}{c}{NB} &
    \multicolumn{1}{c}{Uv} &
    \multicolumn{1}{c}{w/o $\mathbf{G}$} &
    \multicolumn{1}{c|}{ours} & 
    \multicolumn{1}{c}{AN} &
    \multicolumn{1}{c}{NB} &
    \multicolumn{1}{c}{Uv} &
    \multicolumn{1}{c}{w/o $\mathbf{G}$} &
    \multicolumn{1}{c|}{ours} & 
    \multicolumn{1}{c}{AN} &
    \multicolumn{1}{c}{NB} &
    \multicolumn{1}{c}{Uv} &
    \multicolumn{1}{c}{w/o $\mathbf{G}$} &
    \multicolumn{1}{c}{ours} \\ \hline
Dress-Flower & Seq1&
    27.26 &
    \cellcolor{red!25}29.56 &
    27.31 &
    28.25 &
    \cellcolor{orange!25}28.76 &
    0.0492 &
    0.0318 &
    0.029 &
    \cellcolor{orange!25}0.0231 &
    \cellcolor{red!25}0.0217 &
    0.1389 &
    0.0928 &
    0.1002 &
    \cellcolor{orange!25}0.0824 &
    \cellcolor{red!25}0.0783 \\
    \multicolumn{1}{l|}{} & Seq2&
    26.15 &
    \cellcolor{red!25}28.62 &
    26.75 &
    26.25 &
    \cellcolor{orange!25}27.62 &
    0.0575 &
    0.0382 &
    0.0324 &
    \cellcolor{orange!25}0.03 &
    \cellcolor{red!25}0.0247 &
    0.1643 &
    0.1047 &
    0.1065 &
    \cellcolor{orange!25}0.1006 &
    \cellcolor{red!25}0.0851 \\
    \multicolumn{1}{l|}{} & Seq3&
    26.06 &
    \cellcolor{red!25}28.62 &
    26.14 &
    26.5 &
    \cellcolor{orange!25}27.29 &
    0.0614 &
    0.0393 &
    0.0355 &
    \cellcolor{orange!25}0.0298 &
    \cellcolor{red!25}0.0263 &
    0.1744 &
    0.1091 &
    0.1154 &
    \cellcolor{orange!25}0.0975 &
    \cellcolor{red!25}0.0865 \\
    \multicolumn{1}{l|}{} & Seq4&
    25.39 &
    \cellcolor{red!25}28.73 &
    26.49 &
    27.01 &
    \cellcolor{orange!25}27.61 &
    0.0631 &
    0.0411 &
    0.0346 &
    \cellcolor{orange!25}0.0282 &
    \cellcolor{red!25}0.0255 &
    0.177 &
    0.1104 &
    0.1094 &
    \cellcolor{orange!25}0.0902 &
    \cellcolor{red!25}0.0838 \\ \hline
Skirt-Beige & Seq1&
    28.05 &
    \cellcolor{red!25}30.36 &
    29.22 &
    29.43 &
    \cellcolor{orange!25}29.5 &
    0.05 &
    0.0346 &
    0.0292 &
    \cellcolor{orange!25}0.0264 &
    \cellcolor{red!25}0.026 &
    0.1363 &
    0.0884 &
    0.0829 &
    \cellcolor{orange!25}0.0820 &
    \cellcolor{red!25}0.0832 \\
    \multicolumn{1}{l|}{} & Seq2&
    27.38 &
    \cellcolor{red!25}30.22 &
    \cellcolor{orange!25}28.64 &
    28.5 &
    \cellcolor{orange!25}28.64 &
    0.0583 &
    0.0392 &
    0.0329 &
    \cellcolor{orange!25}0.0284 &
    \cellcolor{red!25}0.0286 &
    0.1681 &
    0.1031 &
    0.099 &
    \cellcolor{orange!25}0.0978 &
    \cellcolor{red!25}0.0961 \\ \hline
Skirt-Brown & Seq1&
    25.9 &
    \cellcolor{red!25}26.26 &
    \cellcolor{orange!25}25.56 &
    25.92 &
    25.54 &
    0.0627 &
    0.0458 &
    0.0362 &
    \cellcolor{orange!25}0.0317 &
    \cellcolor{red!25}0.0315 &
    0.1596 &
    0.1153 &
    0.1049 &
    \cellcolor{orange!25}0.0924 &
    \cellcolor{red!25}0.0929 \\
    \multicolumn{1}{l|}{} & Seq2&
    26.29 &
    \cellcolor{red!25}26.91 &
    25.41 &
    \cellcolor{orange!25}25.85 &
    25.83 &
    0.0588 &
    0.0488 &
    0.0359 &
    \cellcolor{orange!25}0.0331 &
    \cellcolor{red!25}0.0305 &
    0.1589 &
    0.1321 &
    0.1069 &
    \cellcolor{orange!25}0.0976 &
    \cellcolor{red!25}0.0952 \\ \hline
Skirt-Cow & Seq1&
    25.78 &
    \cellcolor{red!25}26.7 &
    25.52 &
    \cellcolor{orange!25}26.06 &
    25.58 &
    0.0653 &
    0.0541 &
    0.0393 &
    \cellcolor{orange!25}0.0356 &
    \cellcolor{red!25}0.0338 &
    0.1626 &
    0.1315 &
    0.106 &
    \cellcolor{orange!25}0.0963 &
    \cellcolor{red!25}0.0982 \\ \hline
Skirt-Texture & Seq1&
    28.44 &
    \cellcolor{red!25}29.84 &
    27.9 &
    28.96 &
    \cellcolor{orange!25}29.17 &
    0.0565 &
    0.044 &
    0.0323 &
    \cellcolor{orange!25}0.0255 &
    \cellcolor{red!25}0.0244 &
    0.1546 &
    0.1201 &
    0.0935 &
    \cellcolor{orange!25}0.0818 &
    \cellcolor{red!25}0.0788 \\
    \multicolumn{1}{l|}{} & Seq2&
    28.43 &
    \cellcolor{red!25}30.17 &
    28.25 &
    28.84 &
    \cellcolor{orange!25}29.11 &
    0.0551 &
    0.0396 &
    0.0307 &
    \cellcolor{orange!25}0.0261 &
    \cellcolor{red!25}0.0244 &
    0.1615 &
    0.1149 &
    0.0954 &
    \cellcolor{orange!25}0.0844 &
    \cellcolor{red!25}0.0803 \\ \hline
\end{tabular}%
}
\label{tab:nv_per_seq}
\end{table*}

\begin{table*}[t!]
\caption{Quantitative results of novel pose synthesis. Generally, our method outperforms baselines on PSNR, LPIPS \cite{zhang2018unreasonable}, and DISTS \cite{ding2020image}.
We color cells that have the \colorbox{red!25}{best} and \colorbox{orange!25}{second best} scores.
(AN: Anim-Nerf~\cite{peng2021animatable}, NB: NeuralBody~\cite{peng2021neural}, Uv: UV-NeRF~\cite{chen2023uv} and w/o $\mathbf{G}$: without using garment pose.)}
\resizebox{\textwidth}{!}{%
\begin{tabular}{c|c|lllll|lllll|lllll}
    \hline
    \multirow{2}{*}{Garment} &
    \multirow{2}{*}{Sequence} &
    \multicolumn{5}{c|}{$\mathrm{PSNR}\uparrow$} &
    \multicolumn{5}{c|}{$\mathrm{LPIPS}\downarrow$ \cite{zhang2018unreasonable}} &
    \multicolumn{5}{c}{$\mathrm{DISTS}\downarrow$ \cite{ding2020image}} \\
    & &
    \multicolumn{1}{c}{AN} &
    \multicolumn{1}{c}{NB} &
    \multicolumn{1}{c}{Uv} &
    \multicolumn{1}{c}{w/o $\mathbf{G}$} &
    \multicolumn{1}{c|}{ours} &
    \multicolumn{1}{c}{AN} &
    \multicolumn{1}{c}{NB} &
    \multicolumn{1}{c}{Uv} &
    \multicolumn{1}{c}{w/o $\mathbf{G}$} &
    \multicolumn{1}{c|}{ours} &
    \multicolumn{1}{c}{AN} &
    \multicolumn{1}{c}{NB} &
    \multicolumn{1}{c}{Uv} &
    \multicolumn{1}{c}{w/o $\mathbf{G}$} &
    \multicolumn{1}{c}{ours} \\ \hline
\multicolumn{1}{c|}{Dress-Flower} & Seq1 &
    24.15 &
    \cellcolor{orange!25} 24.53 &
    23.85 &
    24.17 &
    \cellcolor{red!25}25.36 & 
    0.0609 &
    0.0553 &
    0.0517 &
    \cellcolor{orange!25}0.0426 &
    \cellcolor{red!25}0.0387 & 
    0.1734 &
    0.1488 &
    0.1605 &
    \cellcolor{orange!25}0.1155 &
    \cellcolor{red!25}0.1103  \\
    & Seq2 &
    24.76 &
    25.00 &
    24.56 &
    \cellcolor{orange!25} 25.11 &
    \cellcolor{red!25}26.18 &
    0.0642 &
    0.0534 &
    0.0501 &
    \cellcolor{orange!25}0.0385 &
    \cellcolor{red!25}0.034 &
    0.1760 &
    0.1354 &
    0.1616 &
    \cellcolor{orange!25}0.1156 &
    \cellcolor{red!25}0.1088 \\
    & Seq3 &
    \cellcolor{orange!25}24.88 &
    24.85 &
    24.3 &
    25.4 &
    \cellcolor{red!25}26.11 &
    0.0642 &
    0.0518 &
    0.0483 &
    \cellcolor{orange!25}0.037 &
    \cellcolor{red!25}0.0337 &
    0.1800 &
    0.1448 &
    0.1495 &
    \cellcolor{orange!25}0.1119 &
    \cellcolor{red!25}0.1030 \\
    & Seq4 &
    24.63 &
    25.21 &
    24.08 &
    24.94 &
    \cellcolor{red!25}25.85 &
    0.0653 &
    0.053 &
    0.0483 &
    \cellcolor{orange!25}0.038 &
    \cellcolor{red!25}0.0358 &
    0.1851 &
    0.1436 &
    0.1458 &
    \cellcolor{orange!25}0.1097 &
    \cellcolor{red!25} 0.1051 \\ \hline
\multicolumn{1}{c|}{Skirt-Beige} & Seq1&
    25.37 &
    \cellcolor{orange!25} 26.56 &
    25.66 &
    26.52 &
    \cellcolor{red!25}26.58 &
    0.9383 &
    0.9415 &
    0.9377 &
    \cellcolor{orange!25}0.9445 &
    \cellcolor{red!25}0.8968 &
    0.1609 &
    0.1500 &
    0.1672 &
    \cellcolor{orange!25}0.1127 &
    \cellcolor{red!25}0.0840 \\
    & Seq2 &
    25.9 &
    26.58 &
    25.85 &
    \cellcolor{orange!25}26.92 &
    \cellcolor{red!25}27.71 &
    0.0604 &
    0.053 &
    0.0479 &
    \cellcolor{orange!25}0.0403 &
    \cellcolor{red!25}0.0364 &
    0.1733 &
    0.1392 &
    0.1542 &
    \cellcolor{red!25}0.1221 &
    \cellcolor{orange!25}0.1222 \\ \hline
\multicolumn{1}{c|}{Skirt-Brown} &  Seq1&
    24.1 &
    \cellcolor{red!25}24.71 &
    24.00 &
    24.24 &
    \cellcolor{orange!25}24.38 &
    0.0632 &
    0.0533 &
    0.0469 &
    \cellcolor{orange!25}0.0418 &
    \cellcolor{red!25}0.0373 &
    0.1642 &
    0.1405 &
    0.1369 &
    \cellcolor{orange!25}0.1154 &
    \cellcolor{red!25}0.1101 \\
    & Seq2 &
    23.64 &
    \cellcolor{red!25}24.43 &
    23.25 &
    \cellcolor{orange!25}23.95 &
    23.82 &
    0.0701 &
    0.0609 &
    0.053 &
    \cellcolor{orange!25}0.0456 &
    \cellcolor{red!25}0.0419 &
    0.1778 &
    0.1522 &
    0.1467 &
    \cellcolor{orange!25}0.118 &
    \cellcolor{red!25} 0.1137 \\ \hline
\multicolumn{1}{c|}{Skirt-Texture} &  Seq1&
    26.92 &
    \cellcolor{orange!25}27.63 &
    26.52 &
    27.29 &
    \cellcolor{red!25}27.91 &
    0.0576 &
    0.0501 &
    0.0417 &
    \cellcolor{orange!25}0.0342 &
    \cellcolor{red!25}0.0314 &
    0.1628 &
    0.1384 &
    0.1272 &
    \cellcolor{orange!25}0.1025 &
    \cellcolor{red!25}0.0946 \\
    & Seq2 &
    26.43 &
    \cellcolor{red!25}27.9 &
    26.53 &
    27.23 &
    \cellcolor{orange!25}27.46 &
    0.063 &
    0.0489 &
    0.0415 &
    \cellcolor{orange!25}0.0359 &
    \cellcolor{red!25}0.0351 &
    0.1714 &
    0.1292 &
    0.1106 &
    \cellcolor{red!25}0.0988 &
    \cellcolor{orange!25}0.0991 \\ \hline
\end{tabular}%
}
\label{tab:np_per_seq}
\end{table*}

\mparagraph{Number of Views.} All experiments in our main paper are conducted, setting the number of views $N^{v}$ as 8.
Here, we further decrease the number of views $N^{v}$ to 4 and present qualitative results in \reffig{fig:numcams}.
As shown in \reffig{fig:numcams}, our method still produces reasonable fitting results consistent with 2D observations, indicating the robustness of our method.
However, for complex garment geometries like ``Dress-Flower'', our method can cause unnatural mesh distortions when capturing complex motions like dress whirling.

\subsection{Pose-driven Rendering Network}
In this subsection, we evaluate the performance of our method for novel view synthesis on ``Dress-Flower-Seq3''. 

\mparagraph{Metrics.} 
We use PSNR, LPIPS \cite{zhang2018unreasonable} and DISTS \cite{ding2020image} to evaluate the results.
PSNR evaluates the difference between predicted and ground-truth images at pixel level while LPIPS \cite{zhang2018unreasonable} measures their distance at feature level and agrees well with human visual perception.
Moreover, DISTS \cite{ding2020image} measures the structure and texture similarity between the predicted images and the ground truth.

\mparagraph{Motion Weights Initialization.} We compare different initialization strategies mentioned in \refsec{sec:imp_de}. 
As shown in \reftab{tab:prior}, using the blend weights prior from garment T-pose mesh vertices instead of the virtual bones will improve the synthesized image qualities.

\mparagraph{Number of Bones.} All experiments in our main paper are conducted with the number of garment bones $B$ as 25.
Here, we show results with different numbers of bones (10/15/20/25/30) in \reftab{tab:jnum}.
In this setting, both fitting and rendering are conducted using the same number of joints.
We can observe that $B$=25 achieves the best results in terms of all scores, further validating our choice.

\begin{table}[h]
    \caption{Novel view synthesis results of our models trained without and with motion weights initialization using garment T-pose mesh vertices.}
    \centering
    \scalebox{1}{
        \begin{tabular}{c|c|c|c}
            \hline
            & $\mathrm{PSNR}\uparrow$  
            & $\mathrm{LPIPS}\downarrow$ 
            & $\mathrm{DISTS}\downarrow$ \\ \hline
            w/o vertices prior & 26.86 & 0.0272 & 0.0906 \\
            w/ vertices prior  & 27.29 & 0.0263 & 0.0865 \\ \hline
    \end{tabular}}
    \label{tab:prior}
\end{table}

\begin{table}[h]
    \caption{Novel view synthesis results of our models trained with different numbers of garment bones $B$.}
    \centering
    \scalebox{1}{
        \begin{tabular}{c|c|c|c}
            \hline
            Number of Bones  & $\mathrm{PSNR}\uparrow$     
            & $\mathrm{LPIPS}\downarrow$ 
            & $\mathrm{DISTS}\downarrow$ \\ \hline
             $B$=10  & 27.03  & 0.0274   & 0.0912  \\ 
             $B$=15 & 26.95 & 0.0286  & 0.0935 \\
             $B$=20  & 26.96 & 0.0279   & 0.0923 \\
             $B$=25 & 27.29 & 0.0263  &  0.0865  \\
             $B$=30 & 27.12 & 0.0275  & 0.090 \\ \hline
    \end{tabular}}
    \label{tab:jnum}
\end{table}

\begin{table}[h]
    \caption{Novel view synthesis results of our models trained with different numbers of camera views $N^v$.}
    \centering
    \scalebox{0.95}{
        \begin{tabular}{c|c|c|c}
        \hline
            Cameras      & $\mathrm{PSNR}\uparrow$  
            & $\mathrm{LPIPS}\downarrow$   
            & $\mathrm{DISTS}\downarrow$ \\ \hline
            4 & 27.02 & 0.0273 & 0.0882 \\
            8 & 27.29 & 0.0263 & 0.0865 \\ \hline
    \end{tabular}}
    \label{tab:view}
\end{table}

\mparagraph{Number of Views.} 
All experiments in our main paper are conducted with the number of views $N^{v}$ as 8. 
Here, we show results using 4 training views in \reftab{tab:view} (both fitting and rendering are conducted using 4 cameras).
This experiment demonstrates that our method works well on sparse views even with only 4 cameras.

\subsection{Comparisons on Per-sequence}
In our main paper, we report the average scores of our method in terms of PSNR, LPIPS \cite{zhang2018unreasonable} and DISTS \cite{ding2020image} across the entire dataset. 
Here, we provide qualitative comparisons of our method against baselines for each sequence in both novel view and novel pose synthesis, as detailed in \reftab{tab:nv_per_seq} and \reftab{tab:np_per_seq} respectively. 
For brevity, we abbreviate Ani-NeRF \cite{peng2021animatable} as AN, NeuralBody \cite{peng2021neural} as NB, UV-Volumes \cite{chen2023uv} as UV in these tables.
For novel view synthesis, our method demonstrates marginally lower PSNR values compared to NeuralBody\cite{peng2021neural}, yet surpasses all baselines in LPIPS scores. 
This suggests that our method yields results agreeing more with human visual perception. 
Additionally, our method achieves the lowest DISTS scores, implying that our results are more similar to ground-truth images in terms of global structures and local textures.
For novel pose synthesis, our method outperforms all baselines in terms of PSNR, LPIPS and DISTS, showing its superior capability in novel pose generalization.

\end{document}